\pdfoutput=1

\documentclass[11pt]{article}

\usepackage[]{ACL2023}

\usepackage{times}
\usepackage{latexsym}

\usepackage[T1]{fontenc}

\usepackage[utf8]{inputenc}

\usepackage{microtype}

\usepackage{inconsolata}

\usepackage{graphicx}
\usepackage{xcolor}
\usepackage{CJKutf8}
\usepackage{tabularx}
\usepackage{booktabs}
\usepackage{multirow}
\usepackage{float}

\usepackage{array}

\newcolumntype{L}[1]{>{\raggedright\let\newline\\\arraybackslash\hspace{0pt}}m{#1}}
\newcolumntype{C}[1]{>{\centering\let\newline\\\arraybackslash\hspace{0pt}}m{#1}}
\newcolumntype{R}[1]{>{\raggedleft\let\newline\\\arraybackslash\hspace{0pt}}m{#1}}

\title{How Many Answers Should I Give? \\An Empirical Study of Multi-Answer Reading Comprehension}

\author{
    Chen Zhang$^{1}$, 
    Jiuheng Lin$^{1}$, 
    Xiao Liu$^{1}$, 
    Yuxuan Lai$^{3}$, \\
    {\bf Yansong Feng$^{1,2}$\thanks{\;\;Corresponding author.}, 
    Dongyan Zhao$^{1,4,5}$} \\
    $^{1}$ Wangxuan Institute of Computer Technology, Peking University, China\\
    $^{2}$ The MOE Key Laboratory of Computational Linguistics, Peking University, China\\
    $^{3}$ Department of Computer Science, The Open University of China \\
    $^{4}$ State Key Laboratory of Media Convergence Production Technology and Systems \\
    $^{5}$ Beijing Institute for General Artificial Intelligence \\
    {\tt \{zhangch,lxlisa,fengyansong,zhaody\}@pku.edu.cn} \\
    {\tt linjiuheng@stu.pku.edu.cn}~~
    {\tt laiyx@ochn.edu.cn} \\
}

\begin{document}
\maketitle
\begin{abstract}
The multi-answer phenomenon, where a question may have multiple answers scattered in the document, can be well handled by humans but is challenging enough for machine reading comprehension (MRC) systems.
Despite recent progress in multi-answer MRC, there lacks a systematic analysis of how this phenomenon arises and how to better address it.
In this work, we design a taxonomy to categorize commonly-seen multi-answer MRC instances, with which we inspect three multi-answer datasets and analyze where the multi-answer challenge comes from. 
We further analyze how well different paradigms of current multi-answer MRC models deal with different types of multi-answer instances. 
We find that some paradigms capture well the key information in the questions while others better model the relationship between questions and contexts.
We thus explore strategies to make the best of the strengths of different paradigms. 
Experiments show that generation models can be a promising platform to incorporate different paradigms.
Our annotations and code
are released for further research\footnote{\url{https://github.com/luciusssss/how-many-answers}}.
\end{abstract}

\section{Introduction}

\begin{figure}[t]
\centering
\includegraphics[scale=0.4]{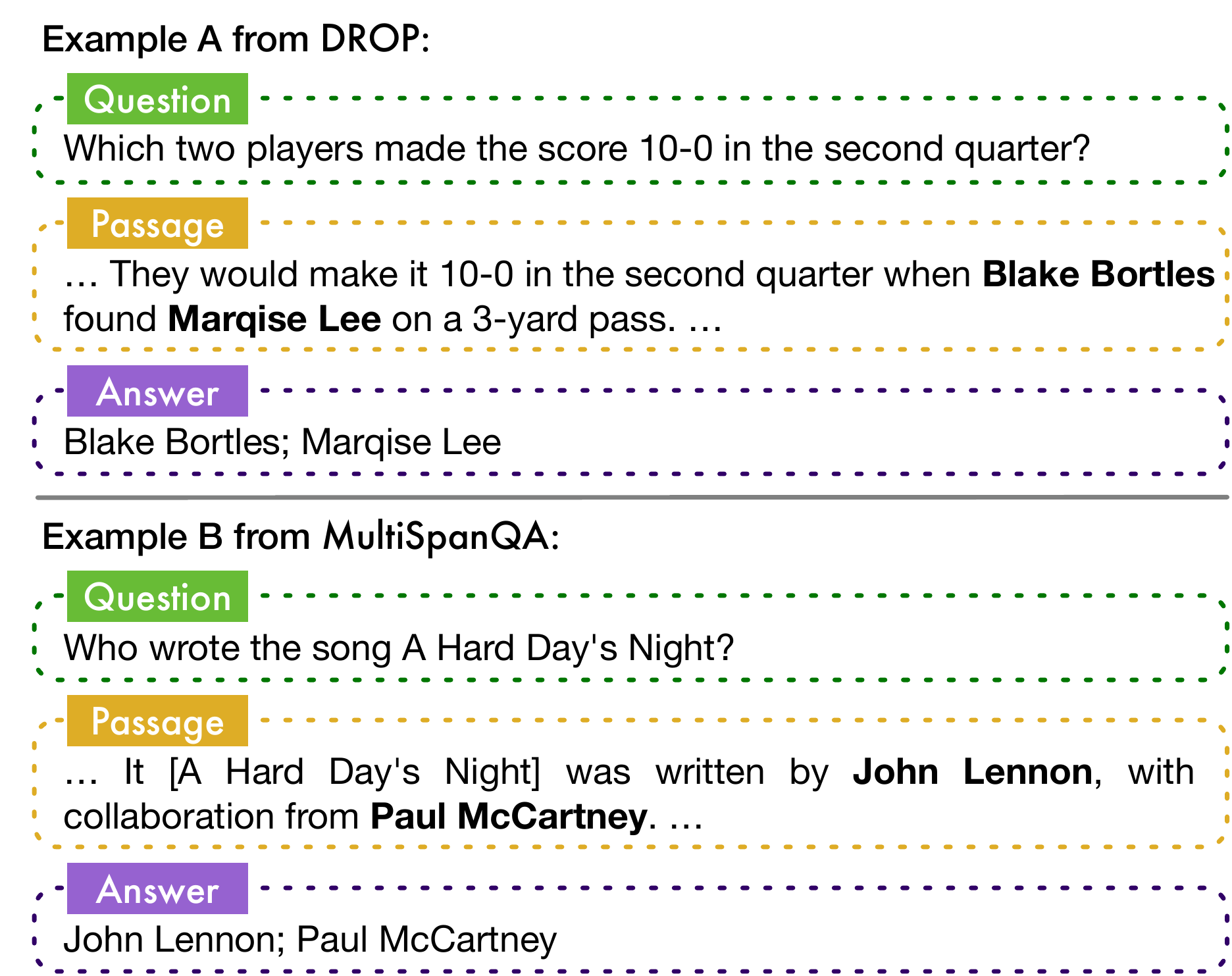}
\caption{Two examples from existing multi-answer MRC datasets. }
\label{fig:first_page_example}
\end{figure}

In the typical setting of machine reading comprehension, such as SQuAD~\cite{rajpurkar-etal-2016-squad}, the system is expected to extract a single answer from the passage for a given question. 
However, in many scenarios, questions may have multiple answers scattered in the passages, and all the answers should be found to completely answer the questions, such as the examples illustrated in Figure~\ref{fig:first_page_example}.
Recently, a series of MRC benchmarks featuring multi-answer instances have been constructed, including DROP~\cite{dua-etal-2019-drop}, Quoref~\cite{dasigi-etal-2019-quoref} and MultiSpanQA~\cite{li-etal-2022-multispanqa}.
Most current research efforts focus primarily on improving the overall QA performance on these benchmarks~\cite{hu-etal-2019-multi,segal-etal-2020-simple,li-etal-2022-multispanqa}. 
Yet, as far as we know, there still lacks a systematic analysis of how the phenomenon of multi-answer arises and how we can better tackle this challenge.

In this paper, we systematically analyze the categorization of multi-answer MRC instances and investigate how to design a strong multi-answer MRC system. 
We try to answer the following research questions: 
(1) Where does the multi-answer challenge come from?
(2) How do different MRC models specifically deal with the multi-answer challenge?
(3) How can we design better models by combining different multi-answer MRC paradigms?

We first analyze existing multi-answer MRC datasets to track the origin of the multi-answer challenge.
Previous works have attempted to categorize multi-answer instances primarily based on the distances or relationships between multiple answers~\cite{li-etal-2022-multispanqa, ju-etal-2022-cmqa}. 
Yet, they did not holistically consider the interaction between questions and contexts.
We observe that in some cases the number of answers is indicated in the question itself (\emph{two players} in Example A of Figure~\ref{fig:first_page_example}) while in others we have no idea until we read the documents carefully (Example B of Figure~\ref{fig:first_page_example}).  

To better understand this challenge,
we develop a taxonomy for the multi-answer phenomenon, based on how the number of answers is determined: the question itself suffices, or both the question and the passage should be taken into consideration.
We annotate 6,857 instances from DROP, Quoref, and MultiSpanQA based on our taxonomy and find that the procedure of dataset construction has a large influence on the expressions in the questions.
Most questions in crowdsourced datasets contain certain clues indicating the number of answers.
By contrast, real-world information-seeking questions are less likely to specify the number of answers, which is usually dependent on the passages.

We further use our annotations to examine the performance of current MRC solutions regarding the multi-answer challenge~\cite{hu-etal-2019-multi,segal-etal-2020-simple,li-etal-2022-multispanqa}, which can be categorized into 4 paradigms, i.e., \textsc{Tagging}, \textsc{NumPred}, \textsc{Iterative} and \textsc{Generation}.
We analyze their strengths and weaknesses and find that some efforts, e.g., \textsc{NumPred}, are good at capturing the key information in the questions, while others, e.g., \textsc{Iterative}, can better model the relation between questions and contexts. 
This motivates us to investigate better ways to benefit from different paradigms.

Given the complementary nature of these paradigms, we wonder whether a combination of paradigms improves performance on multi-answer MRC.
We explore two strategies, early fusion and late ensemble, to benefit from different paradigms.
With a generation model as the backbone, we attempt to integrate the paradigms \textsc{NumPred} and \textsc{Interative}, in a lightweight Chain-of-Thought style~\cite{wei2022chain}. 
Experiments show that the integration remarkably improves the performance of generation models, demonstrating that \textsc{Generation} is a promising platform for paradigm fusion.

Our contributions are summarized as follows: 
(1) We design a taxonomy for multi-answer MRC instances according to how the number of answers can be determined. 
It considers both questions and contexts simultaneously, enlightening where the multi-answer challenge comes from.
(2) We annotate 6,857 instances from 3 datasets with our taxonomy, which enables us to examine 4 paradigms for multi-answer MRC in terms of their strengths and weaknesses.
(3) We explore various strategies to benefit from different paradigms. Experiments show that generation models are promising to be backbones for paradigm fusion.

\section{Task Formulation}
\label{sec:task_formulation}
In multi-answer MRC, given a question $Q$ and a passage $P$, a model should extract several spans, $A = \{ a_1, a_2, ..., a_n \} (n\geq 1)$, from $P$  to answer $Q$. 
Each span, $a_i \in A$, corresponds to a partial answer to $Q$, and the answer set $A$ as a whole answers $Q$ completely.
These spans can be contiguous or discontiguous in the passage. 

We distinguish between two terms, \textit{multi-answer} and \textit{multi-span}, which are often confused in previous works.
\textit{Multi-answer} indicates that a question should be answered with the complete set of entities or utterances.
\textit{Multi-span} is a definition from the perspective of answer annotations. 
In certain cases, the answer annotation of a question can be either single-span or multi-span, as explained in the next paragraph.
Ideally, we expect that the answers to a multi-answer question should be annotated as multi-span in the passage, where each answer is grounded to a single span, although some of them can be contiguous in the passage. 
\advance\leftmargini -1em 
\begin{quote}
\small
    \textbf{Q0}: What's Canada's official language? \\
    \textbf{P}: [...] \textbf{English} and \textbf{French}, are the official languages of the Government of Canada. [...] \\
    \vspace{-\baselineskip}
\end{quote}

For example, in Q0, there are two answers, \textit{English} and \textit{French}, to the given question.
According to the annotation guidelines of SQuAD, one might annotate this instance with a single continuous span \textit{English and French}.
Yet, this form of annotation is not preferred in the multi-answer MRC setting. It blurs the boundary of different answers and fails to denote explicitly the number of expected answers. 
Thus, it is suboptimal for a comprehensive model evaluation. 
Instead, we suggest denoting each answer with distinct spans, say, annotating this instance with two spans, \textit{English} and \textit{French}.
With this criterion, we can encourage models to disentangle different answers. 
With fine-grained answer annotations, we can also assess how well a model answers a question sufficiently and precisely.

This annotation criterion generally conforms to the annotation guidelines of existing multi-answer datasets, e.g., DROP, Quoref and MultiSpanQA. 
A few instances violating the criterion are considered as bad annotations, as discussed in Section~\ref{sec:annotation_proccess}. See more remarks on the task formulation in Appendix~\ref{app:task_formulation}.

\section{Taxonomy of Multi-Answer MRC} 
\label{sec:taxonomy}
\begin{figure}[t]
\centering
\includegraphics[scale=0.2]{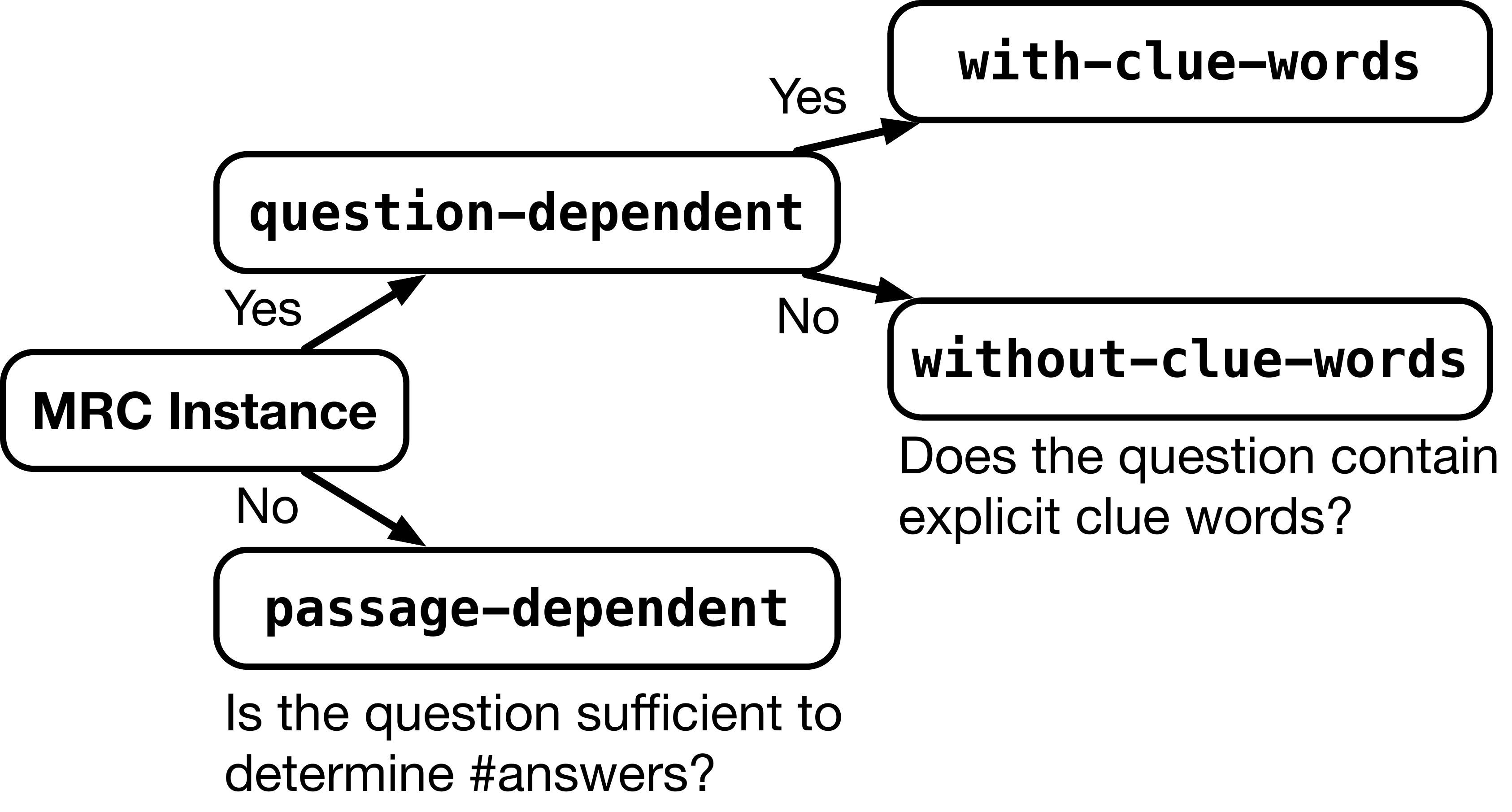}
\caption{Illustration of our taxonomy for multi-answer MRC instances. }
\label{fig:taxonomy}
\end{figure}

To better understand the challenge of multi-answer, we first design a taxonomy to categorize various multi-answer MRC instances. 
It assesses how the number of answers relates to the question or passage provided. 
Different from the previous works that classify questions according to the distances or relations between multiple answers~\cite{li-etal-2022-multispanqa, ju-etal-2022-cmqa}, 
our taxonomy, taking both questions and passages into consideration, focuses on how the number of answers is determined. 
This enables us to analyze multi-answer questions and single-answer questions in a unified way.
We illustrate our taxonomy in Figure~\ref{fig:taxonomy} and elaborate on each category as follows. 

\begin{table}[t]
\small
\centering
\begin{tabular}{L{0.23\columnwidth}L{0.45\columnwidth}C{0.11\columnwidth}}
\toprule
\textbf{Type} & \textbf{Question} & \textbf{\# Ans.} \\
\midrule
Cardinal & Which \textbf{two} players completed 1-yard TD pass? & 2\\
\midrule
Ordinal & Who scored the \textbf{first} touchdown of the game? & 1 \\
\midrule
Comp./Super. & What's the \textbf{largest} pizza chain in America? & 1\\
\midrule
Alternative & Is San Juan Bautista incorporated \textbf{or} unincorporated? & 1\\
\midrule
Other Semantics & What are the first names of the \textbf{trio} who try to call 911? & 3\\
\bottomrule
\end{tabular}
\caption{Examples of various types of clue words. Comp./Super. denotes comparatives and superlatives.}
\label{tab:clue_word_example}
\end{table}

\paragraph{\texttt{Question-Dependent}} If one can infer the exact number of answers from the question without referring to the passage, this instance belongs to the \texttt{question-dependent} category.
According to whether there are clue words that directly indicate the number of answers, this type is further divided into two sub-categories:

(a) In a \texttt{with-clue-words} question, one can find a few words that indicate the number of answers. In Q1, the word \textit{two} in the question indicates that two answers are expected. 
\begin{quote} 
\small
    \textbf{Q1}: What are the two official languages of Puerto Rico? \\
    \textbf{P}: [...] \textbf{English} is an official language of the Government of Puerto Rico. [...] As another official language, \textbf{Spanish} is widely used in Puerto Rico. [...] \\
    \vspace{-\baselineskip}
\end{quote}
We group the clue words into five types: cardinal, ordinal, comparative/superlative, alternative, and other lexical semantics, as illustrated in Table~\ref{tab:clue_word_example}. 

(b) In a \texttt{without-clue-words} question, although we can not locate obvious clue words, we can infer the number of answers with sentence semantics or commonsense knowledge.
In Q2, we can determine that there is only one conversion result for the question based on sentence semantics instead of any single words. 
\begin{quote}
\small
    \textbf{Q2}: 1 light year equal to how many km?\\
    \textbf{P}: [...] The light-year is a unit of length used to express astronomical distances. It is about \textbf{9.5 trillion kilometres} or 5.9 trillion miles. [...] \\
    \vspace{-\baselineskip}
\end{quote}
In Q3, we can infer that the following question has only one answer, based on the commonsense that there is only one winner of a given Super Bowl.
\begin{quote}
\small
    \textbf{Q3}: Who won Super Bowl XXXIX? \\
    \textbf{P}: [...] The Eagles advanced to Super Bowl XXXIX, where they dueled the 2004 \textbf{New England Patriots} season. [...] The Patriots won 24-21. [...] \\
    \vspace{-\baselineskip}
\end{quote}

\paragraph{\texttt{Passage-Dependent}}
In a \texttt{passage-dependent} instance, the question itself is not adequate to infer the number of answers. One needs to rely on the provided passage to decide how many answers are needed to answer the question. In Q4, we have no idea of the number of answers solely based on the question. If we refer to the passage, we will find ten answers to the question. 

\begin{quote}
\small
    \textbf{Q4}: Which countries does the Danube River flow through? \\
    \textbf{P}: [...] Originating in \textbf{Germany}, the Danube flows southeast for 2,850 km, passing through or bordering \textbf{Austria}, \textbf{Slovakia}, \textbf{Hungary}, \textbf{Croatia}, \textbf{Serbia}, \textbf{Romania}, \textbf{Bulgaria}, \textbf{Moldova} and \textbf{Ukraine} before draining into the Black Sea. [...]\\
    \vspace{-\baselineskip}
\end{quote}

\section{Analyses of Multi-Answer Datasets}
We investigate existing multi-answer datasets based on our designed taxonomy to analyze where the multi-answer challenge comes from. 

\subsection{Datasets}
\label{sec:source_datasets}
We annotate the validation sets of three widely-used multi-answer MRC datasets, i.e., DROP~\cite{dua-etal-2019-drop}, Quoref~\cite{dasigi-etal-2019-quoref}, and MultiSpanQA~\cite{li-etal-2022-multispanqa}. 
The number of annotated questions is listed in Table~\ref{tab:number_of_questions} and more statistics are in Appendix~\ref{appendix:data_stats}.

\textbf{DROP} 
 is a crowdsourced MRC dataset for evaluating the discrete reasoning ability.
The annotators are encouraged to devise questions that require discrete reasoning such as arithmetic.
DROP has four answer types: numbers, dates, single spans, and sets of spans. 
Since the previous two types of answers are not always exact spans in the passages, we only consider the instances whose answers are single spans or sets of spans.

\textbf{Quoref}  focuses on the coreferential phenomena. 
The questions are designed to require resolving coreference among entities.  
10\% of its instances require multiple answer spans.

\textbf{MultiSpanQA} is a dataset specialized for multi-span reading comprehension. The questions are extracted from NaturalQuestions~\cite{kwiatkowski-etal-2019-natural}, which are real queries from the Google search engine.

\begin{table}
\centering
\small
\setlength{\tabcolsep}{7pt} 
\begin{tabular}{lcrr}
\toprule
\textbf{Dataset}  & 
\textbf{All} & 
\textbf{Single-Ans.} & \textbf{Multi-Ans.}\\
\midrule
DROP & 3,133 & 2,609 & 524 \\
Quoref & 2,418 & 2,198 & 220 \\
MultiSpanQA & 1,306 & 653 & 653 \\
\midrule
Total & 6,857 & 5,460 & 1,397\\
\bottomrule
\end{tabular}
\caption{The number of instances for human annotation in the validation set of each dataset.}
\label{tab:number_of_questions}
\end{table}

\subsection{Annotation}
\label{sec:annotation_proccess}
\paragraph{Annotation Process}
Our annotation process is two-staged: we first automatically identify some \texttt{question-dependent} instances and then recruit annotators to classify the remaining ones.

In the first stage, we automatically identify the questions containing certain common clue words such as numerals (full list in Appendix~\ref{appendix:annotation}) to reduce the workload of whole-process annotation. 
Afterward, the annotators manually check whether each instance is \texttt{question-dependent}. Out of the 4,594 recalled instances, 3,727 are identified as \texttt{question-dependent}.

In the second stage, we recruit annotators to annotate the remaining 3,130 instances. 
For each instance, given both the question and the answers, the annotators should first check whether the form of answers is correct and mark incorrect cases as \texttt{bad-annotation}\footnote{In the first stage, the annotators also need to check whether an instance is \texttt{bad-annotation}. }. 
We show examples of common \texttt{bad-annotation} cases in Table~\ref{tab:bad_annotation}.
After filtering out the \texttt{bad-annotation} ones, the annotators are presented with the question only and should decide whether they could determine the number of answers solely based on the question. If so, this instance is annotated as \texttt{question-dependent}; otherwise \texttt{passage-dependent}.
For a \texttt{question-dependent} instance,
the annotators are further asked to extract the clue words, if any, from the question, which determines whether the instance is \texttt{with-clue-words} or \texttt{without-clue-words}.

\paragraph{Quality Control}
Six annotators participated in the annotation after qualification.
Each instance is annotated by two annotators.
In case of any conflict, a third annotator resolves it.
An instance is classified as \texttt{bad-annotation} if any annotator labels it as \texttt{bad-annotation}.
Cohen's Kappa between two initial annotators is 0.70, indicating substantial agreement. See more details in Appendix~\ref{appendix:annotation}.

\subsection{Analyses of Annotation Results}

\begin{table*}
\centering
\small
\setlength{\tabcolsep}{7pt} 
\begin{tabular}{l|r|rrr|r}
\toprule
\textbf{Dataset}  & \textbf{\texttt{passage-dependent}} & \multicolumn{3}{c|}{\textbf{\texttt{question-dependent}}}  & \textbf{\texttt{bad-annotation}} \\
 & & \multicolumn{1}{c}{All} & \texttt{with-clue-word} & \texttt{no-clue-word} & \\
\midrule
DROP & 826 (26.4\%) & \textbf{2,242 (71.6\%)} & 2,204 (70.3\%) & 38~~~(1.2\%) & 65 (2.1\%) \\
Quoref & 711 (29.4\%) & \textbf{1,704 (70.5\%)} & 1,639 (67.8\%) & 65~~~(2.7\%) & 3 (0.2\%) \\
MultiSpanQA & \textbf{991 (75.9\%)} & 285 (21.8\%) & 121~~~(9.3\%) & 164 (12.6\%) & 30 (2.3\%) \\
\midrule
Total & 2,528 (36.9\%) & 4,231 (61.7\%) & 3,964 (57.8\%) & 267~~~(3.9\%) & 98 (1.4\%)\\
\bottomrule
\end{tabular}
\caption{Distribution of instance types in three datasets.}
\label{tab:main_annotation_results}
\end{table*}

\begin{table*}
\centering
\small
\setlength{\tabcolsep}{5pt} 
\begin{tabular}{l|c|rrrrr}
\toprule
\textbf{Dataset}  & \textbf{\texttt{with-clue-word}} &  \textbf{Cardinal} & \textbf{Ordinal} & \textbf{Comp./Super.} & \textbf{Alternative} & \textbf{Other Semantics} \\
\midrule
DROP & 2,204 & 113~~~~(5.1\%) & 592 (26.9\%) & \textbf{1,298 (58.9\%)} & 1,214 (55.1\%) & 135~~~ (6.1\%) \\
Quoref & 1,639 & 83~~~~(5.1\%) & 35~~~(2.1\%) &  25~~~~(1.5\%) & 0~~~(0.0\%) & \textbf{1,501 (91.6\%)} \\
MultiSpanQA & ~~~121 & \textbf{51 (41.8\%)} & 26~(21.3\%) & 23~~(19.0\%) & 2~~~(1.6\%) & 19~~(15.6\%)\\
\bottomrule
\end{tabular}
\caption{Distribution of clue word types in three datasets. A question may contain multiple types of clue words.}
\label{tab:substype_annotation_results}
\end{table*}

With our annotated data, we study how the multi-answer instances differ across different datasets under our designed taxonomy. We find that the distributions of instance types are closely related to how the datasets are constructed. 

\paragraph{Instance Types}
The distributions of instance types in different datasets are shown in Table~\ref{tab:main_annotation_results}. 
\texttt{Question-dependent} prevails in DROP and Quoref, making up over 70\% of the two datasets. 
In contrast, most instances in MultiSpanQA are \texttt{passage-dependent}. 
This difference stems from how the questions are collected.

DROP and Quoref use crowdsourcing to collect questions with specific challenges.
Given a passage, the annotators know the answers in advance and produce questions that can only be answered through certain reasoning skills.
These artificial questions are more likely to contain clues to the number of answers, such as the question with ordinal in Table~\ref{tab:clue_word_example}.
By contrast, the questions in MultiSpanQA are collected from search engine queries. 
Users generally have no idea of the answers to the queries. The number of answers, as a result, is more often dependent on the provided passages, such as Q4 in Section~\ref{sec:taxonomy}.

\paragraph{Clue Words}
Since a large portion (57.8\%) of the annotated instances belong to the \texttt{with-clue-word} type, we further investigate the distribution of clue words in different datasets, shown in Table~\ref{tab:substype_annotation_results}.
On the one hand, the questions contain a large variety of clue words, demonstrating the complexity of multi-answer MRC.
On the other hand, the prevailing type of clue words is different in each dataset, reflecting the preference in dataset construction.
Specifically, nearly 60\% of the \texttt{with-clue-word} questions in DROP are alternative questions with comparatives/superlatives, because DROP's annotators are encouraged to inject discrete reasoning challenges, e.g., comparison, when writing questions.
In Quoref, 91\% of the clue words indicate the number of answers through their lexical semantics. 
This unbalanced distribution results from the emphasis on coreference resolution: most questions begin with \textit{what is the name of the person who ...}, where \textit{name of the person} is identified as clue words. 
In MultiSpanQA, whose questions are search engine queries, 63\% of the \texttt{with-clue-word} questions contain numerals. 
If users already know the number of desired answers, they tend to restrict it in the question, such as \textit{seven wonders of the world}.

We provide more analyses on of how the instance types are distributed with respect to the specific number of answers in Appendix~\ref{app:additional_analyses_annotation}.

\section{Existing Multi-Answer MRC Models}
\label{sec:describe_paradigm}

\begin{figure*}[ht]
\centering
\includegraphics[scale=0.2]{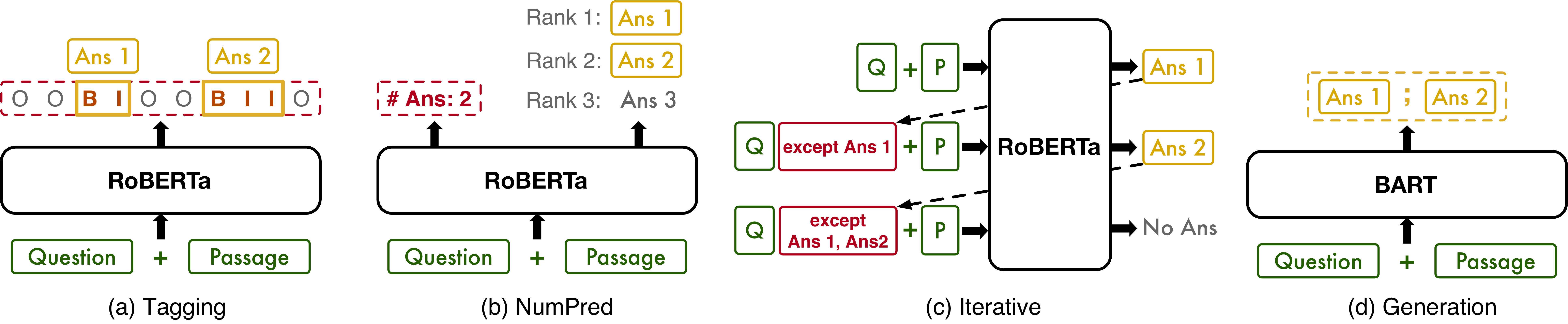}
\caption{An illustration of four paradigms for multi-answer MRC. } 
\label{fig:illustration_of_paradigm}
\end{figure*}

Based on our categorization of the multi-answer instances, we continue to investigate how existing multi-answer MRC models perform differently on various types of multi-answer instances. 
We summarize current solutions into four paradigms according to how they obtain multiple answers, as illustrated in  Figure~\ref{fig:illustration_of_paradigm}.

\paragraph{\textsc{Tagging}}
\citet{segal-etal-2020-simple} cast the multi-answer MRC task as a sequence tagging problem, similar to named entity recognition (NER), so that the model can extract multiple non-contiguous spans from the context. 

\paragraph{\textsc{NumPred }(Number Prediction)}
\citet{hu-etal-2019-multi} first predict the number of answers $k$ as an auxiliary task and then select the top $k$ non-overlapped ones from the output candidate spans.

\paragraph{\textsc{Iterative}}
Searching for evidence iteratively is widely adopted in many QA tasks~\cite{xu-etal-2019-enhancing,zhao-etal-2021-multi-step,zhang-etal-2021-extract-integrate}, but it is not explored in multi-answer MRC. 
We adapt this idea to extract multiple answers iteratively.
In each iteration, we append the previously extracted answers to the question, with the word \textit{except} in between, and then feed the updated question to a single-answer MRC model.
The iterative process terminates when the model predicts no more answers.

\paragraph{\textsc{Generation}}
Generation has been adopted as a uniform paradigm for many QA tasks \cite{khashabi-etal-2020-unifiedqa, khashabi2022unifiedqa}, but it is less explored on multi-answer MRC.
For \textsc{Generation}, we concatenate all answers, with semicolons as separators, to form an output sequence, and finetune the model to generate it conditioned on the question and passage.

\subsection{Experimental Setup}
\paragraph{Implementation Details}
We use RoBERTa-base~\cite{liu2019roberta} for the three extractive paradigms and BART-base~\cite{lewis-etal-2020-bart} for \textsc{Generation}.
We train models on the training sets of each dataset and evaluate them on the corresponding validation sets with our instance type annotations.
See more details in Appendix~\ref{appendix:implementation}.

\paragraph{Metrics}
We adopt the official metrics of MultiSpanQA~\cite{li-etal-2022-multispanqa}, including the precision (P), recall (R), and F1 in terms of exact match (EM) and partial match (PM).
See Appendix~\ref{appendix:metrics} for details.

\begin{table}
\centering
\small
\setlength{\tabcolsep}{3.5pt} 
\begin{tabular}{l|ccc|ccc}
\toprule
\textbf{Model} & \multicolumn{3}{c|}{\textbf{EM}} & \multicolumn{3}{c}{\textbf{PM}} \\
 & \textbf{P} & \textbf{R} & \textbf{F1} & \textbf{P} & \textbf{R} & \textbf{F1}\\
\midrule
\multicolumn{7}{c}{{DROP}} \\ 
\midrule
\textsc{Tagging} & \textbf{61.86} &	\textbf{63.91} &	\textbf{62.87} & \textbf{77.53} &	\textbf{77.39} &	\textbf{77.46} \\ 
\textsc{NumPred} & 61.59 &	56.77 &	59.09 &	76.71 &	74.86 &	75.77 \\
\textsc{Iterative} & 60.66 &	60.07 &	60.36 &	76.19 &	76.04 &	76.11 \\
\textsc{Generation} &	60.07 &	57.15 &	58.58 &	75.39 &	72.39 &	73.86 \\
\midrule
\multicolumn{7}{c}{{Quoref}} \\ 
\midrule
\textsc{Tagging}	& \textbf{71.00} &	\textbf{72.21} &	\textbf{71.60} &	\textbf{80.44} &	\textbf{79.74} &	\textbf{80.09} \\
\textsc{NumPred} & 65.61 &	63.57 &	64.57 &	77.30 &	78.20 	& 77.75 \\
\textsc{Iterative} &	67.28 &	66.35 &	66.81 &	78.57 &	78.58 &	78.57 \\
\textsc{Generation} &	63.57 &	63.39 &	63.48 &	73.38 &	74.02 &	73.70 \\
\midrule
\multicolumn{7}{c}{{MultiSpanQA}} \\ 
\midrule
\textsc{Tagging} &	61.31 &	\textbf{68.84} &	64.85 &	80.45 &	\textbf{83.08} &	81.75 \\
\textsc{NumPred} &	55.03 &	46.06 &	50.15 &	80.16 &	75.26 &	77.63 \\
\textsc{Iterative} &	\textbf{66.32} &	67.98 &	\textbf{67.14} &	\textbf{84.39} &	80.96 &	\textbf{82.64} \\
\textsc{Generation}	& 65.40 & 62.60 &	63.97 &	82.06 &	78.14 &	80.06 \\
\bottomrule
\end{tabular}
\caption{Performance of four paradigms on three datasets.}
\label{tab:overall_experiment_results}
\end{table}

\begin{table}
\centering
\small
\setlength{\tabcolsep}{5pt}
\begin{tabular}{l|c|ccc}
\toprule
\textbf{Model}  & \textbf{\texttt{p-dep.}} & \multicolumn{3}{c}{\textbf{\texttt{q-dep.}}}  \\
 & & ~~~~All~~~~ & \texttt{w/-clue} & \texttt{w/o-clue} \\
\midrule
\multicolumn{5}{c}{{DROP}} \\ 
\midrule
\textsc{Tagging}	& \textbf{74.57}	& \textbf{79.11}	& \textbf{80.88}	& 68.77\\
\textsc{NumPred} & 72.37 &	77.54 &	79.32 &	70.08 \\
\textsc{Iterative} & 73.47	& 77.60 & 79.21	& 65.73 \\
\textsc{Generation}	& 72.18 &	74.77	& 76.19 & \textbf{72.62} \\
\midrule
\multicolumn{5}{c}{{Quoref}} \\ 
\midrule
\textsc{Tagging} & 70.60 &	\textbf{84.86}	& \textbf{85.23} &	75.76 \\
\textsc{NumPred} & 69.45 &	81.88 &	82.44 &	70.12 \\
\textsc{Iterative} &	\textbf{71.42} &	82.18 &	82.37 &	\textbf{77.30} \\
\textsc{Genration}	& 66.31	& 77.41	& 78.38 &	52.63\\
\midrule
\multicolumn{5}{c}{{MultiSpanQA}} \\ 
\midrule
\textsc{Tagging} &	82.28 &	79.66 &	86.60 &	73.36 \\
\textsc{NumPred} & 77.77 &	77.11 &	78.19 &	\textbf{78.77} \\
\textsc{Iterative} &	\textbf{82.78} &	\textbf{82.09} &	\textbf{87.22} &	77.80 \\
\textsc{Generation} &	80.57 &	78.05 &	81.73 &	75.85 \\
\bottomrule
\end{tabular}
\caption{The performance (PM F1) of four paradigms on different types of instances.
\texttt{p-dep.} denotes \texttt{passage-dependent}.
\texttt{q-dep.} denotes \texttt{question-dependent}.}
\label{tab:subtype_experiment_results}
\end{table}

\subsection{Results and Analyses} 
We report the overall performance in Table~\ref{tab:overall_experiment_results}, and the performance on different instance types in Table~\ref{tab:subtype_experiment_results}. 
We observe that each of these paradigms has its own strengths and weaknesses.

\textsc{Tagging} outperforms other paradigms on DROP and Quoref, whose dominating instance type is \texttt{question-dependent}. 
Although \textsc{Tagging} has no explicit answer number prediction step, it can still exploit this information implicitly because it takes the question into account during the sequential processing of every token.
Besides, \textsc{Tagging}, as a common practice for entity recognition, is good at capturing the boundaries of entities. 
Thus, it is not surprising that it performs the best on DROP and Quoref, most of whose answers are short entities.

\textsc{Iterative} achieves the best overall performance on MultiSpanQA, whose prevailing instance type is \texttt{passage-depenent}. This paradigm does not directly exploit the information of the number of answers given in the question. 
Rather, it encourages adequate interactions between questions and passages, performing single-answer extraction at each step. As a result, \textsc{Iterative} does well for the questions whose number of answers heavily depends on the given context.

As for \textsc{NumPred}, although we expect high performance on \texttt{question-dependent} instances, it lags behind \textsc{Tagging} by approximately 2\% in PM F1 on DROP and Quoref. 
This might result from the gap between training and inference.
The model treats the answer number prediction and answer span extraction as two separate tasks during training, with limited interaction. Yet during inference, the predicted number of answers is used as a hard restriction on multi-span selection.
Different from the decent performance on DROP and Quoref, \textsc{NumPred} performs worst among the four paradigms on MultiSpanQA, because it is difficult for models to accurately predict the number of answers for a long input text that requires thorough understanding. 

Among all paradigms, \textsc{Generation} generally performs the worst. 
Under the same parameter scale, extractive models seem to be the better choice for tasks whose outputs are exact entity spans from the input, while generation models do well in slightly longer answers.
This also explains the smaller gap between \textsc{Generation} and extractive paradigms on MultiSpanQA compared to that on DROP and Quoref: 
MultiSpanQA has many descriptive long answers instead of short entities only.

\begin{figure}[t]
\centering
\includegraphics[scale=0.38]{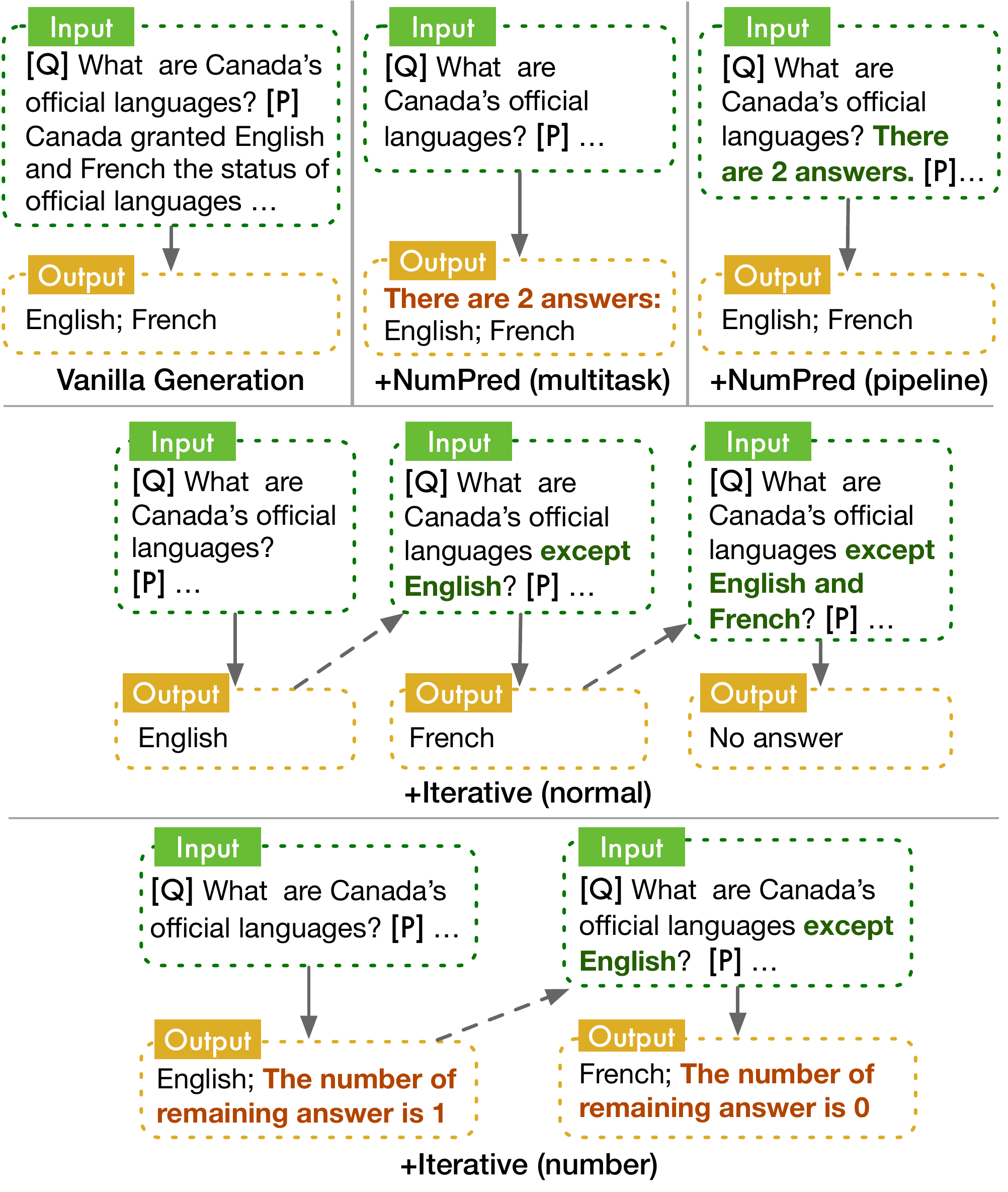}
\caption{An illustration of different strategies for early fusion of paradigms.}
\label{fig:fusion}
\end{figure}
\section{Fusion of Different Paradigms}

From the above analysis, we can see that extractive methods can better locate exact short spans in the passage, and \textsc{NumPred} can provide potential guidance on the number of answers. Meanwhile, the generation models can better handle longer answers and are more adaptable to different forms of inputs and outputs. 
Now an interesting question is how to combine different paradigms to get the best of both worlds.

We explore two strategies for combining different paradigms: \textbf{early fusion} and \textbf{late ensemble}. 
The former mixes multiple paradigms in terms of model architectures while the latter ensembles the predictions of different models.
We discuss our exploration of late ensemble in Appendix~\ref{appendix:late_ensemble} since model ensemble is a well-explored technique.
Here we primarily elaborate on early fusion.
We carry out a series of pilot studies to demonstrate the potential of paradigm fusion.

Previous works attempt to fuse two extractive paradigms, \textsc{Tagging} and \textsc{NumPred}~\cite{segal-etal-2020-simple,li-etal-2022-multispanqa}. However, they only lead to marginal improvements, probably because \textsc{Tagging} can already implicitly determine answer numbers well and the help of \textsc{NumPred} is thus limited. 

Although the performance of base-size generation models on multi-answer MRC is inferior to that of extractive ones, generation models of larger sizes show great potential with more parameters and larger pre-training corpora~\cite{khashabi-etal-2020-unifiedqa, khashabi2022unifiedqa}. 
More importantly, \textsc{Generation} can easily adapt to various forms of inputs and outputs.
We carry out pilot studies using a generation model as the backbone and benefiting from the ideas of other paradigms. 
We propose several lightweight methods to combine \textsc{Generation} with \textsc{NumPred} and \textsc{Iterative}, as illustrated in Figure~\ref{fig:fusion}.

\paragraph{\textsc{Generation} + \textsc{NumPred}}
Inspired by recent works on Chain-of-Thought \cite{wei2022chain}, we guide the  model with prompts indicating the number of answers.
We introduce a \textbf{\textsc{NumPred} prompt sentence} (NPS) in the form of \textit{There are $\{2, 3, ...\}$ answers/There is only one answer}. 
We experiment with two variants, multitask and pipeline.
In the multitask variant, the model outputs an NPS before enumerating all the answers. 
In the pipeline variant, we predict the number of answers with a separate classifier and then append the NPS to the question as extra guidance.

\paragraph{\textsc{Generation} + \textsc{Iterative}}
We substitute the original extractor of \textsc{Iterative} with a generator.
The iterative process terminates when the model outputs the string \textit{No answer}. 
Besides the normal setting, we experiment with another variant that additionally outputs an NPS in the form of \textit{The number of remaining answers is $\{1, 2, 3, ...\}$}.

\paragraph{Results}
Our main experiments are conducted with BART-base and BART-large due to our limited computational budget.
For the pipeline variant of \textsc{Generation} + \textsc{NumPred}, we use RoBERTa-base as an answer number classifier.
The overall experiment results are reported in Table~\ref{tab:fusion_experiment_results} and the results on different question types are reported in Appendix~\ref{appendix:early_fusion}.

When \textsc{Generation} is multitasking with \textsc{NumPred}, it outperforms the vanilla one consistently.
The NPS in the output provides a soft but useful hint for the succeeding answer generation, improving the accuracy of answer number prediction by 1.7\% on average for BART-base. 
The pipeline variant is often inferior to the multitasking one due to error propagation. 
Especially, its performance drops a lot on MultiSpanQA, whose instances are \texttt{passage-dependent}. The accuracy of the answer number classifier on MultiSpanQA lags behind that on the other two datasets by more than 12\%. Thus the NPS in the input, with an unreliably predicted answer number, is more likely to mislead the subsequent answer span generation.

The combination of \textsc{Generation} and \textsc{Iterative} does not always lead to improvement.
This might be because the answer generation process of \textsc{Generation} is already in an iterative style: in the output sequence, each answer is generated conditioned on the previously-generated ones.
The incorporation of \textsc{Iterative} thus does not lead to further improvement.
When we further introduce an NPS with the number of remaining answers, the performance generally outperforms the normal setting.
This proves that \textsc{Generation}, as a backbone, is easy to integrate with various hints.

\begin{table}[ht]
\centering
\small
\setlength{\tabcolsep}{5pt} 
\begin{tabular}{l|cc|cc}
\toprule
\multirow{2}*{\textbf{Model}} & \multicolumn{2}{c|}{\textbf{Base}} & \multicolumn{2}{c}{\textbf{Large}} \\
 & \textbf{EM} & \textbf{PM} & \textbf{EM} & \textbf{PM}\\
\midrule
\multicolumn{5}{c}{{DROP}} \\ 
\midrule
Vanilla \textsc{Generation} & 58.58 & 73.86 & 66.43 & 80.55 \\
+\textsc{NumPred} (multitask) & \textbf{60.02} & \textbf{74.34} & \textbf{69.61} & \textbf{82.85} \\
+\textsc{NumPred} (pipeline) & 59.19 & 73.94 & 66.45 & 80.63 \\
+\textsc{Iterative} (normal) & 58.44 & 73.58 & 66.55 & 80.53 \\
+\textsc{Iterative} (number) & 58.98 & 74.07 & 68.19 & 82.17 \\
\midrule
\multicolumn{5}{c}{{Quoref}} \\ 
\midrule
Vanilla \textsc{Generation} & 63.48  & 73.70 & 76.57 & 84.47 \\
+\textsc{NumPred} (multitask) & 66.25 & 75.43 & \textbf{77.04} & 84.45  \\
+\textsc{NumPred} (pipeline) & 67.94 & 77.42 & 75.42 & 83.66 \\
+\textsc{Iterative} (normal) & \textbf{68.81} & \textbf{78.23} & 74.72 & 82.60 \\
+\textsc{Iterative} (number) & 63.33 & 73.34 & 76.67 & \textbf{84.57} \\
\midrule
\multicolumn{5}{c}{{MultiSpanQA}} \\ 
\midrule
Vanilla \textsc{Generation} & 63.97  & 80.06 & 69.13 & 84.61  \\
+\textsc{NumPred} (multitask) & \textbf{64.85} & \textbf{80.58} & \textbf{69.31} & \textbf{84.82} \\
+\textsc{NumPred} (pipeline) & 39.71 & 60.94 & 45.34 & 68.09 \\
+\textsc{Iterative} (normal) & 63.26 & 79.97 & 65.62 & 82.88 \\
+\textsc{Iterative} (number) & 63.84 & 80.04 & 66.77 & 83.41 \\
\bottomrule
\end{tabular}
\caption{The performance (EM F1 and PM F1) of different strategies for early fusion of paradigms. }
\label{tab:fusion_experiment_results}
\end{table}

\begin{table}[ht]
\centering
\small
\setlength{\tabcolsep}{5pt} 
\begin{tabular}{l|c|cc}
\toprule
\textbf{Model} & \textbf{Setting} & \textbf{EM F1} & \textbf{PM F1} \\
\midrule
Vanilla BART-base & Supervised & 66.77 & 81.24 \\
Vanilla BART-large & Supervised & 71.93 & 85.83 \\
Vanilla GPT-3.5 & One-Shot & 53.34 & 79.27 \\
GPT-3.5 + \textsc{NumPed} & One-Shot & 63.45 & 82.38 \\
\bottomrule
\end{tabular}
\caption{The performance of BART and GPT-3.5 on the multi-answer instances of MultiSpanQA. }
\label{tab:chatgpt_results}
\end{table}

\paragraph{Pilot Study on GPT-3.5} 
To investigate whether these fusion strategies work on larger models, we conduct a pilot study on GPT-3.5. 
We use the 653 multi-answer instances in the validation set of MultiSpanQA for experiments. The prompts are listed in Appendix~\ref{appendix:early_fusion}. The experiment results are shown in Table~\ref{tab:chatgpt_results}. 

When given only one example for in-context learning, GPT-3.5 can already achieve 79.27\% PM F1 on the multi-answer instances, with only a small gap between BART trained on full data. 
Its EM F1 score is low because GPT-3.5 cannot handle the boundaries of answer spans well. This is not unsurprising since one example is not sufficient for GPT-3.5 to learn the annotation preference of span boundaries in MultiSpanQA. 
If we ask GPT-3.5 to predict the number of answers before giving all the answers, we observe an improvement of 10.1\% EM F1 and 3.1\% PM F1. This proves the effectiveness of fusing \textsc{NumPed} with larger generation models 

As evidenced by the above trials, it is promising to fusion different paradigms.
We hope that our exploration will inspire future works adopting larger generation models for multi-answer MRC.

\section{Related Works}
Compared to the vast amount of single-answer MRC datasets, the resources for multi-answer MRC are limited. 
Aside from the datasets in Section~\ref{sec:source_datasets}, MASH-QA~\cite{zhu-etal-2020-question} focuses on the healthcare domain, with 27\% of the questions having multiple long answers, ranging from phrases to sentences.
CMQA~\cite{ju-etal-2022-cmqa} is another multi-answer dataset in Chinese, featuring answers with conditions or different granularities. 
For our analysis, we select 
two commonly-used datasets, DROP and Quoref, as well as a newly-released dataset, MultiSpanQA.

Current models addressing multi-answer MRC generally fall into two paradigms: \textsc{Tagging}~\cite{segal-etal-2020-simple} and \textsc{NumPred}~\cite{hu-etal-2019-multi}, as explained in Section~\ref{sec:describe_paradigm}.
\textsc{Iterative}~\cite{xu-etal-2019-enhancing,zhao-etal-2021-multi-step,zhang-etal-2021-extract-integrate,gao-etal-2021-answering} and \textsc{Generation}~\cite{khashabi-etal-2020-unifiedqa, khashabi2022unifiedqa} have been adopted for many types of QA tasks including knowledge base QA, multiple-choice QA, and open-domain QA. 
Nevertheless, their performance on multi-answer MRC is less explored. In our paper, we also study how to adapt these paradigms for multi-answer MRC. 
Apart from the exploration of model architectures for multi-answer MRC, \citet{lee2023liquid} attempt to generate multi-answer questions as data augmentation. 

Previous works have made preliminary attempts in  fusing two extractive paradigms.
\citet{segal-etal-2020-simple} adopt a single-span extraction model for single-answer questions and \textsc{Tagging} for multi-answer questions; 
\citet{li-etal-2022-multispanqa} add a \textsc{NumPred} head to the \textsc{Tagging} framework. The predicted number of answers is used to adjust the tagging results. 
Both strategies lead to marginal improvement over the baselines.
We instead resort to \textsc{Generation} for paradigm fusion, considering its potential with larger sizes and its flexibility in inputs and outputs.

\section{Conclusion}
In this paper, we conduct a systematic analysis for multi-answer MRC. 
We design a new taxonomy for multi-answer instances based on how the number of answers is determined.
We annotate three datasets with the taxonomy and find that multi-answer is not merely a linguistic phenomenon; rather, many factors contribute to it, especially the process of data collection.
With the annotation, we further investigate the performance of four paradigms for multi-answer MRC and find their strengths and weaknesses.
This motivates us to explore various strategies of paradigm fusion to boost performance. 
We believe that our taxonomy can help determine what types of questions are desirable in the annotation process and aid in designing  more practical annotation guidelines.
We hope that our annotations can be used for more fine-grained diagnoses of MRC systems and encourage more robust MRC models.

\section*{Limitations}
First, our taxonomy of multi-answer MRC instances only considers whether we know the \textit{exact} number of answers from the questions. 
In some cases, one might have an \textit{imprecise estimate} of answer numbers from the question. 
For example, for the question \textit{Who are Barcelona's active players?}, one might estimate that there are dozens of active players for this football club.
Yet, these estimations are sometimes subjective and difficult to quantify.
Therefore, this instance is classified as \texttt{passage-dependent} according to our current taxonomy.
We will consider refining our taxonomy to deal with these cases in the future.

Second, we did not conduct many experiments with pre-trained models larger than the large-size ones due to limited computational budgets. 
Generation models of larger sizes show great potential with more parameters and larger pre-training corpora.
We encourage more efforts to deal with multi-answer MRC with much larger models, such as GPT-3.5.

\section*{Acknowledgments}
This work is supported by NSFC~(62161160339).
We would like to thank the anonymous reviewers for their valuable suggestions, and our great annotators for their careful work, especially Zhenwei An, Nan Hu, and Hejing Cao.
Also, we would like to thank Quzhe Huang for his help in this work. 
For any correspondence, please contact Yansong Feng.

\bibliography{anthology,custom}
\bibliographystyle{acl_natbib}

\clearpage
\appendix
\section{Additional Remarks on Task Formulation}
\label{app:task_formulation}
As discussed in Section~\ref{sec:task_formulation}, \textit{multi-answer} and \textit{multi-span} are two orthogonal concepts. We have already shown an example (Q0 in Section~\ref{sec:task_formulation}) where a \textit{multi-answer} question can be annotated as \textit{single-span} by certain annotation guidelines. 
Here is another example to demonstrate the difference between \textit{multi-answer} and \textit{multi-span}.

\begin{quote}
\small
    \textbf{Q}: Which offer of Triangle-Transit is most used by students? \\
    \textbf{P}: [...] Triangle-Transit offers \textbf{scheduled}, fixed-route regional and commuter \textbf{bus service}. The first is most used by students. \\
    \vspace{-\baselineskip}
\end{quote}

This is an example where a \textit{single-answer} question can be annotated as \textit{multi-span}.
A single answer, \textit{scheduled bus service}, will be annotated as multiple-span, i.e., \textit{scheduled} and \textit{bus service} in the passage. 

Considering the differences between \textit{multi-answer} and \textit{multi-span}, we suggest carefully distinguishing between these two terms in the future.

\section{Annotation Details}
\label{appendix:annotation}

\paragraph{Dataset Statistics}
We report more statistics of the annotated datasets in Table~\ref{tab:dataset_statistics}.
MultiSpanQA has the largest average number of answers since it is a dataset  designed especially for multi-answer questions. The answers in MultiSpanQA are generally longer than those in DROP and Quoref because many of the answers in MultiSpanQA are long descriptive phrases or clauses instead of short entities. 
For all three datasets, the distances between answers are large.
This indicates that the answers to a large proportion of the questions are discontiguous in the passages, demonstrating the difficulty of multi-answer MRC. 

\label{appendix:data_stats}
\begin{table}[h]
    \centering
    \small
    \resizebox{\linewidth}{!}{
    \begin{tabular}{l|ccc}
      \toprule
      \textbf{Dataset} & \textbf{DROP} & \textbf{Quoref} & \textbf{MultiSpanQA} \\
      \midrule
      Length of Question  & 9.4 & 15.5 & 9.0\\
      Length of Context  & 214.7 & 326.0 & 219.9\\
      Length of Answer  & 1.9 & 1.6 & 3.1\\
      \midrule
      \#Answers  & 1.2 & 1.1 & 1.9\\
      \#Answers (Multi) & 2.5 & 2.4 & 2.9\\
      \midrule
      Distance Between Ans. & 30.5 & 17.3 & 10.3\\
      \bottomrule
    \end{tabular}
    }
    \caption{Dataset Statistics, including the (a) average length (in words) of questions, contexts, and answers, (b) the average number of answers for all the instances and the multi-answer ones, (c) the average distances (in words) between answers. }
    \label{tab:dataset_statistics}
\end{table}

\paragraph{Pre-defined Clue Words}
Here, we list the  pre-defined clue words in the first stage of annotation: 
    \vspace{-\topsep}
\begin{itemize}
    \vspace{-\topsep}
    \item Numerals, including cardinals and ordinals
        \vspace{-\topsep}
    \item Comparatives and superlatives
        \vspace{-\topsep}
    \item The word \textit{or}, as an indicator of alternative questions.
        \vspace{-\topsep}
    \item Other words, including \textit{only}, \textit{last}, \textit{single}, \textit{name of the person}, and, \textit{top}.
\end{itemize}

\paragraph{Selection of Annotators}
A total of 10 graduates proficient in English participated in our annotation task. 
We first provided training materials to the annotators and asked them to annotate 100 sample instances.
Based on their annotation accuracy on the sample instances, six of them are qualified to continue annotating the remaining instances.
The annotators are paid \$10 per hour, which is  adequate given the participants’ demographic.
The annotators are informed of how the data would be used. 

\paragraph{Examples of Bad Annotations}
In Table~\ref{tab:bad_annotation}, we present several examples we marked as \texttt{bad-annotation}. Common reasons for bad annotations including incorrect segmentation of answers, irrelevant answers, and duplicate answers.

\begin{table*}
\small
\centering
\begin{tabular}{p{0.4\columnwidth}p{0.75\columnwidth}p{0.55\columnwidth}}
\toprule
\textbf{Type}  & \textbf{Example} & \textbf{Explanation} \\
\midrule
Incorrect segmentation of answers
& \textbf{Source: } DROP

\textbf{Question:} Which event occurred first, Duke Magnus Birgersson started a war or Erik Klipping gathered a large army?

\textbf{Annotated Answers:} Duke Magnus Birgersson; started a war
& The correct answer \textit{Duke Magnus Birgersson started a war} is wrongly split into two spans, 
\textit{Duke Magnus Birgersson} and \textit{started a war}. \\
\midrule

Irrelevant Answers
& \textbf{Source: } DROP

\textbf{Question:} Who scored first in the second half of the game, Cowboys or 49ers?

\textbf{Annotated Answers:} end of the half; San Francisco scored; making the score 28-14
& All three annotated answers are not related to the questions. A correct answer should be either \textit{Cowboys} or \textit{49ers}. \\
\midrule

Duplicate answers 
& \textbf{Source: } MultiSpanQA

\textbf{Question:} who benefited by title ix of the education amendments 

\textbf{Annotated Answers:} women; women playing college sports 
& One annotead answer, \textit{women} is duplicated with the other, \textit{women playing college sports}. \\

\bottomrule
\end{tabular}
\caption{Examples and explanations of \texttt{bad-annotation} cases. }
\label{tab:bad_annotation}
\end{table*}

\section{Additional Analyses on Annotation Results}
\label{app:additional_analyses_annotation}

We report more statistics of the annotation results in Table~\ref{tab:answer_statistics} and Table~\ref{tab:clue_word_statistics}, and conduct additional analyses from the perspective of the number of answers.

For multi-answer instances, \texttt{passage-dependent} questions account for the largest proportion, followed by \texttt{with-clue-word}. 
As for the single-answer instances in DROP and Quoref, they tend to be \texttt{question-dependent}, while in MultiSpanQA most of them are \texttt{passage-dependent}. 
In terms of the clue words in the \texttt{with-clue-word} questions, cardinal numbers are more common in multi-answer questions while other types of clue words are more likely to appear in single-answer questions.

\begin{table}[ht]
\centering
\small
\setlength{\tabcolsep}{12pt}
\resizebox{\linewidth}{!}{
\begin{tabular}{l|ccc}
\toprule
\textbf{\#Ans} & \multicolumn{1}{c|}{\textbf{\texttt{p-dep.}}} & \multicolumn{2}{c}{\textbf{\texttt{q-dep.}}} \\
 & \multicolumn{1}{l|}{} & \texttt{w/-clue} & \texttt{w/o-clue} \\ 
\midrule
\multicolumn{4}{c}{DROP} \\
\midrule
1 & 480 & \textbf{2,085} & 37 \\
2 & \textbf{209} & 105 & 1 \\
3 & \textbf{74} & 11 & 0 \\
>3 & \textbf{63} & 3 & 0 \\ 
\midrule
\multicolumn{4}{c}{Quoref} \\
\midrule
1 & 582 & \textbf{1,548} & 65 \\
2 & \textbf{82} & 62 & 0 \\
3 & \textbf{28} & 23 & 0 \\
>3 & \textbf{19} & 6 & 0 \\ 
\midrule
\multicolumn{4}{c}{MultiSpanQA} \\
\midrule
1 & \textbf{448} & 56 & 140 \\
2 & \textbf{300} & 32 & 22 \\
3 & \textbf{131} & 14 & 2 \\
>3 & \textbf{112} & 19 & 0  \\ 
\bottomrule              
\end{tabular}
}
\caption{Distribution of question types according to the number of answers.
\texttt{p-dep.} denotes \texttt{passage-dependent}.
\texttt{q-dep.} denotes \texttt{question-dependent}.}
\label{tab:answer_statistics}
\end{table}

\begin{table}[h]
\centering
\small
\setlength{\tabcolsep}{2pt} 
\resizebox{\linewidth}{!}{
\begin{tabular}{l|ccccc}
\toprule
\textbf{\#Ans}  & \textbf{Alternative} & \textbf{Cardinal} & \textbf{Comp./Super.} & \textbf{Ordinal} & \textbf{Others} \\ 
\midrule
\multicolumn{6}{c}{{DROP}} \\
\midrule
1 & 1,213 & 3 & 1,293 & 588 & 132 \\
2 & 1 & 97 &  4 & 3 & 3 \\
3 & 0 & 11 & 0 & 0 & 0\\
>3 & 0 & 2 & 1 & 1 & 0\\
\midrule
\multicolumn{6}{c}{{Quoref}} \\
\midrule
1 & 0 & 1 & 25 & 35 & 1,492 \\
2 & 0 & 55 &  0 & 0 & 7 \\
3 & 0 & 21 & 0 & 0 & 2\\
>3 & 0 & 6 & 0 & 0 & 0\\
\midrule
\multicolumn{6}{c}{{MultiSpanQA}} \\
\midrule
1 & 2 & 1 & 14 & 25 & 14 \\
2 & 0 & 21 &  5 & 1 & 5 \\
3 & 0 & 12 & 2 & 0 & 0\\
>3 & 0 & 17 & 2 & 0 & 0\\
\bottomrule
\end{tabular}
}
\caption{Distribution of clue word types in three datasets according to the number of answers.}
\label{tab:clue_word_statistics}
\end{table}

\section{Experimental Setup}
\subsection{Implementation Details}
\label{appendix:implementation}
We use base-size models for our main experiments for sake of energy savings.
Since T5-base has twice as many parameters as RoBERTa-base and BART-base, we did not use it to ensure fair comparisons.
We carefully tune each model on the training set and report its best performance on the validation set. 
We use an NVIDIA A40 GPU for experiments. A training step takes approximately 0.5s for RoBERTa-base and 0.2s for BART-base. 
We describe the implementation details of different models here. 

\paragraph{\textsc{Tagging}} We use the implementation 
 by~\citet{segal-etal-2020-simple}\footnote{\url{https://github.com/eladsegal/tag-based-multi-span-extraction}}. 
 We use the IO tagging variant, which achieves the best overall performance according to the original paper. 
 We adopt the best-performing hyperparameters provided by the original paper. 
 
\paragraph{\textsc{NumPred}} Because the implementation by the original paper~\cite{hu-etal-2019-multi}\footnote{\url{https://github.com/huminghao16/MTMSN}} does not support RoBERTa, we re-implement the model with Huggingface Transformers~\cite{wolf-etal-2020-transformers}\footnote{\url{https://github.com/huggingface/transformers}}. We use the representation of the first token in the input sequence for answer number classification. The maximum number of answers of the classifier is 8. 
The batch size is 12. The number of training epochs  is 10.
The learning rate is 3e-5. 
The maximum sequence length is 512. 

\paragraph{\textsc{Iterative}} 
Our implementation is based on the scripts of MRC implemented by Huggingface. 
During training, the order of answers for each iteration is determined by their order of position in the passage. 
The batch size is 8. The number of training epochs  is 8.
The learning rate is 3e-5. 
The maximum sequence length is 384.
During inference, the beam size is set to 3 and the length penalty is set to 0.7. The maximum length of answers is 10.

\paragraph{\textsc{Generation}}
Our implementation is based on the scripts of sequence generation implemented by Huggingface. The batch size is 12. The learning rate is 3e-5. The number of training epochs is 10.
The maximum input length is 384. 
The maximum output length is 60.

\subsection{Evaluation Metrics}
\label{appendix:metrics}
Here, we describe the evaluation metrics used in our experiments, which are the official ones used by MultiSpanQA~\cite{li-etal-2022-multispanqa}. 
The metrics consist of two part: exact match and partial match. 

\paragraph{Exact Match}
An exact match occurs when a prediction fully matches one of the ground-truth answers.
We use micro-averaged precision, recall, and F1 score for evaluation.

\paragraph{Partial Match} 
For each pair of prediction $p_i$ and ground truth answer $t_j$ , the partial retrieved score $s_{ij}^{ret}$ and partial relevant score $s_{ij}^{rel}$ are calculated as the length of the longest common substring (LCS) between $p_i$ and $t_j$, divided by the length of $p_i$ and $t_j$ respectively, as:
$$s_{ij}^{ret}=\frac{\mathrm{len}(\mathrm{LCS}(p_i,t_j))}{\mathrm{len}(p_i)}$$
$$s_{ij}^{rel}=\frac{\mathrm{len}(\mathrm{LCS}(p_i,t_j))}{\mathrm{len}(t_j)}$$

Suppose there are $n$ predictions and $m$ ground
truth answers for a question.
We compute the partial retrieved score between a prediction and all answers and keep the highest one as the retrieved score of that prediction. 
Similarly, for each ground truth answer, the relevant score is the highest one between it and all predictions. 
The precision, recall, and F1 are finally defined as follows:
$$\mathrm{Precision}=\frac{\sum^n_{i=1}\mathrm{max}_{j\in[1,m]}(s_{ij}^{ret})}{n}$$
$$\mathrm{Recall}=\frac{\sum^m_{j=1}\mathrm{max}_{i\in[1,n]}(s_{ij}^{rel})}{m}$$
$$\mathrm{F1}=\frac{\mathrm{2*Precision*Recall}}{\mathrm{Precision+Recall}}$$
We use micro-averaged scores for these metrics.

\section{Additional Experiment Results}
\subsection{Late Ensemble}
\label{appendix:late_ensemble}
By late ensemble, we aggregate the outputs from models of different paradigms to boost performance. 
We experiment with a simple voting strategy.
If a span is predicted as an answer by more than one model, we add it to the final prediction set. If a span is part of another span, we consider them equivalent and take the longer one.
In rare cases where the four models predict totally different answers, we add them all to the final prediction set.

Our voting strategy leads to improvements of 1.0\%, 1.2\%, and 1.3\% in PM F1 on DROP, Quoref, and MultiSpanQA, respectively, over the best-performing models in Table~\ref{tab:overall_experiment_results}. 
Yet, this strategy might discard many correct answers.
In the future, we can explore more sophisticated strategies. 
For example, similar to the idea of Mixture of Experts~\cite{jacobs1991adaptive}, the system can evaluate the probability that the instance belongs to a certain category and then adjust the weight of the model based on its capabilities in this category.

\subsection{Early Fusion}
\label{appendix:early_fusion}
In Table~\ref{tab:subtype_experiment_results_of_fusion}, we report the performance of different strategies for early fusion on different types of instances. In Table~\ref{tab:chatgpt_prompts}, we list the prompts used for our pilot study on GPT-3.5.

\begin{table*}
\centering
\small
\setlength{\tabcolsep}{5pt}
\begin{tabular}{l|c|ccc|c|ccc}
\toprule
 & \multicolumn{4}{c|}{\textbf{BART-base}} & \multicolumn{4}{c}{\textbf{BART-large}} \\
\cmidrule{2-9}
\textbf{Model}  & \textbf{\texttt{p-dep.}} & \multicolumn{3}{c|}{\textbf{\texttt{q-dep.}}} & \textbf{\texttt{p-dep.}} & \multicolumn{3}{c}{\textbf{\texttt{q-dep.}}} \\
 & & ~~~~All~~~~ & \texttt{w/-clue} & \texttt{w/o-clue} & &~~~~All~~~~ & \texttt{w/-clue} & \texttt{w/o-clue} \\
\midrule
\multicolumn{9}{c}{{DROP}} \\ 
\midrule
Vanilla \textsc{Generation} & 72.18 & 74.77 & 76.19 & 72.62 & 78.57 & 81.65 & 83.42 & 77.31\\
+\textsc{NumPred} (multitask) & \textbf{72.45} & 75.37 & 76.80 & 70.58 & 80.35 & \textbf{84.24} & \textbf{86.06} & 77.88  \\
+\textsc{NumPred} (pipeline) & 70.58 & \textbf{75.72} & \textbf{77.14} & \textbf{76.77} & 76.79 & 82.66 & 84.34 & \textbf{79.65} \\
+\textsc{Iterative} (normal) & 71.82 & 74.55 & 75.97 & 68.26 & 78.07 & 81.90 & 83.56 & 74.03 \\
+\textsc{Iterative} (number) & 71.90 & 75.27 & 76.66 & 72.93 & \textbf{80.58} & 83.05 & 84.91 & 72.66 \\
\midrule
\multicolumn{9}{c}{{Quoref}} \\ 
\midrule
Vanilla \textsc{Generation} & 66.31 & 77.41 & 78.38 & 52.63 & 76.41 & \textbf{88.51} & \textbf{88.90} & 79.76 \\
+\textsc{NumPred} (multitask) & 67.54 & 79.37 & 80.15 & 58.30 & \textbf{77.73} & 87.88 & 88.11 & \textbf{82.35} \\
+\textsc{NumPred} (pipeline) & 66.26 & \textbf{82.88} & \textbf{83.55} & 65.20 & 75.37 & 87.71 & 88.22 & 77.36 \\
+\textsc{Iterative} (normal) & \textbf{69.40} & 82.68 & 83.24 & \textbf{68.24} & 73.13 & 87.43 & 87.93 & 77.20  \\
+\textsc{Iterative} (number) & 65.79 & 77.16 & 77.97 & 55.63 & 77.69 & 88.09 & 88.59 & 75.41 \\
\midrule
\multicolumn{9}{c}{{MultiSpanQA}} \\ 
\midrule
Vanilla \textsc{Generation} & 80.57 & 78.05 &  81.73 & 75.85 & 84.52 & \textbf{84.96} & 88.78 & \textbf{81.65} \\
+\textsc{NumPred} (multitask) & \textbf{81.08} & 78.65 & 81.09 & \textbf{77.73} & \textbf{84.83} & 84.80 & \textbf{89.66} & 81.06\\
+\textsc{NumPred} (pipeline) & 60.24 & 63.56 & 68.67 & 58.25 & 67.27 & 71.21 & 74.53 & 69.33 \\
+\textsc{Iterative} (normal) & 80.46 & 78.06 & 81.81 & 74.78 & 83.16 & 81.78 & 84.84 & 80.87 \\
+\textsc{Iterative} (number) & 80.15 & \textbf{79.63} & \textbf{83.47} & 76.17 & 83.49 & 83.06 & 86.08 & 80.44 \\
\bottomrule
\end{tabular}
\caption{The performance (PM F1) of different strategies for early fusion on different types of instances.
\texttt{p-dep.} denotes \texttt{passage-dependent}.
\texttt{q-dep.} denotes \texttt{question-dependent}.}
\label{tab:subtype_experiment_results_of_fusion}
\end{table*}

\begin{table*}
\centering
\small
\setlength{\tabcolsep}{5pt}
\begin{tabular}{p{2.0\columnwidth}r}
\toprule
\textbf{Vanilla GPT-3.5} \\
Answer the question based on the given context. Each question has more than one answer. Please give all the answers and separate them with a semicolon.\\
Context: Laura Horton is a fictional character from the NBC soap opera , Days of Our Lives , a long - running serial drama about working class life in the fictional , United States town of Salem . Created by writer Peggy Phillips , the role was originated by actress Floy Dean on June 30 , 1966 till October 21 , 1966 . Susan Flannery stepped into the role from November 22 , 1966 to May 27 , 1975 . Susan Oliver briefly stepped into the role from October 10 , 1975 , to June 9 , 1976 , followed by Rosemary Forsyth from August 24 , 1976 , to March 25 , 1980 .\\
Question: who played laura horton on days of our lives\\
Answers: Floy Dean; Susan Flannery; Susan Oliver; Rosemary Forsyth\\
\\
Following the example above and answer the following multi-answer question. Please give all the answers and separate them with a semicolon.\\
Context: \texttt{\{context\}}\\
Question: \texttt{\{question\}}\\
Answers: \\
\midrule
\textbf{GPT-3.5 + \textsc{NumPed}} \\
Answer the question based on the given context. Each question has more than one answer. Please predict the number of answers first, then give all the answers and separate them with a semicolon.\\
Context: Laura Horton is a fictional character from the NBC soap opera , Days of Our Lives , a long - running serial drama about working class life in the fictional , United States town of Salem . Created by writer Peggy Phillips , the role was originated by actress Floy Dean on June 30 , 1966 till October 21 , 1966 . Susan Flannery stepped into the role from November 22 , 1966 to May 27 , 1975 . Susan Oliver briefly stepped into the role from October 10 , 1975 , to June 9 , 1976 , followed by Rosemary Forsyth from August 24 , 1976 , to March 25 , 1980 .\\
Question: who played laura horton on days of our lives\\
Answers: The number of answers is 4: Floy Dean; Susan Flannery; Susan Oliver; Rosemary Forsyth\\
\\
Following the example above and answer the following multi-answer question. Please predict the number of answers first, then give all the answers and separate them with a semicolon.\\
Context: \texttt{\{context\}}\\
Question: \texttt{\{question\}}\\
Answers: \\
\bottomrule
\end{tabular}
\caption{The one-shot prompts for GPT-3.5 to answer multi-answer questions in MultiSpanQA. }
\label{tab:chatgpt_prompts}
\end{table*}

\section{Licenses of Scientific Artifacts}
The license for Quoref and DROP is CC BY 4.0.
The license for HuggingFace Transformers is Apache License 2.0.
Other datasets and models provide no licenses.

\end{document}